\documentclass[letterpaper, 10 pt, conference]{ieeeconf}  

\IEEEoverridecommandlockouts                              

\usepackage{graphicx} 
\graphicspath{ {Figures/} }
\usepackage{color,soul}
\usepackage{amsmath} 
\usepackage{amssymb}  
\usepackage{textcomp} 
\usepackage[noadjust]{cite}
\usepackage{multirow}
\usepackage[hidelinks]{hyperref}
\usepackage{booktabs}

\newcommand\inv[1]{#1\raisebox{1.15ex}{$\scriptscriptstyle-\!1$}}

\title{\LARGE \bf
Autonomous Scanning for Endomicroscopic Mosaicing and 3D Fusion
}

\author{Lin Zhang, Menglong Ye, Petros Giataganas, Michael Hughes and Guang-Zhong Yang
\thanks{All authors are with The Hamlyn Centre for Robotic Surgery, Institute of Global Health Innovation,
	Imperial College London, SW7 2AZ, London, UK}
}

\begin{document}

\maketitle
\thispagestyle{empty}
\pagestyle{empty}

\begin{abstract}
Robotic-assisted Minimally Invasive Surgery (RMIS) can benefit from the automation of common, repetitive or well-defined but ergonomically difficult tasks. One such task is the scanning of a pick-up endomicroscopy probe over a complex, undulating tissue surface in order to enhance the effective field-of-view through video mosaicing. In this paper, the da Vinci\textsuperscript{\textregistered} surgical robot, through the dVRK framework, is used for autonomous scanning and 2D mosaicing over a user-defined region of interest. To achieve the level of precision required for high quality mosaic generation, which relies on sufficient overlap between consecutive image frames, visual servoing is performed using a combination of a tracking marker attached to the probe and the endomicroscopy images themselves. The resulting sub-millimetre accuracy of the probe motion allows for the generation of large endomicroscopy mosaics with minimal intervention from the surgeon. It also allows the probe to be maintained in an orientation perpendicular to the local tissue surface, providing optimal imaging results. Images are streamed from the endomicroscope and overlaid live onto the surgeon’s view, while 2D mosaics are generated in real-time, and fused into a 3D stereo reconstruction of the surgical scene, thus providing intuitive visualisation and fusion of the multi-scale images. The system therefore offers significant potential to enhance surgical procedures, by providing the operator with cellular-scale information over a larger area than could typically be achieved by manual scanning.
\end{abstract}

\section{Introduction} \label{Introduction}

Minimally Invasive Surgery (MIS), in which surgical procedures are performed through small incisions, is a widely used and effective approach for surgical oncology. It has significant advantages over the traditional open techniques, in terms of reduced blood loss and postoperative pain, lower infection rates, and shorter hospital stays. However, MIS also has some drawbacks, which include constrained motion due to `keyhole' access, poor depth perception from only a monocular laparoscope, and lack of direct tissue interaction. To provide better manipulation and depth perception, robotic systems with stereo vision have been developed for MIS. In particular, the da Vinci\textsuperscript{\textregistered} robot (Intuitive Surgical Inc., CA) is a successful surgical platform which has been widely used in the treatment of gynecological and urological cancer. However, the da Vinci\textsuperscript{\textregistered} robot is purely a master-slave system that only provides tele-operation abilities to the surgeons, with no automation or artificial intelligence features.

Whilst human guidance is essential for MIS, a recent review \cite{moustris2011autosurgreview} has concluded that surgical automation can be used for some important steps of the operation, which can then reduce the cognitive load of the surgeon during certain surgical tasks that require repetitive and precise motion. To this end, studies have been conducted, for instance, to maintain consistent motion for ultrasound elastography \cite{billings2012us}, or to automatically compensate heart motion in cardiovascular surgery \cite{ruszkowski2015dvrk}. In \cite{padoy2011humansuture}, a human-machine collaborative framework was proposed to improve surgeons' performance by semi-automating surgical subtasks. Further studies have also investigated how to perform automatic surgical debridement \cite{Murali2015}, tissue dissection \cite{pratt2015autonomous}, and brain ablation \cite{hu2015semibrain}.

Another obvious target for automation is the process of generating \emph{in situ} microscopic images of tissue using probe-based confocal laser endomicroscopy (pCLE), such that the microscopic images can used for \emph{in vivo} and \emph{in situ} pathology analysis. Endomicroscopy has been used extensively for diagnostic and surgical procedures in the gastrointestinal tract and abdominal organs \mbox{\cite{wallace2010preliminary,meining2011pcle}}, mainly to discriminate between normal and cancerous tissue regions. However, the miniaturisation requirement for the probe results in a limited field-of-view (FoV) of typically 0.25-1 mm, and therefore, it is difficult to characterise larger areas of tissue or to return to previously sampled points \cite{Ye2013}. Furthermore, retrieving high quality and stable microscopic images with a manually manipulated probe is a challenging task, particularly when the operator attempts to `mosaic' images to synthesise a larger FoV comparable to histological images. Mosaicing requires the operator to maintain optimal probe pose and tissue-contact while performing slow and controlled scanning motions with sub-millimetre accuracy \cite{Latt2011}.

Recent work has proposed using robotised instruments to enhance control of the endomicroscopy probe, allowing large mosaics to be formed to assist real-time diagnosis. These developments have included novel robotic mechanisms for 2D scanning \cite{rosa2013building,dwyer2015miniaturised,Zuo2015breast} or for maintaining a desired probe-tissue contact force \cite{Latt2011}. Integration with existing robotic systems, such as the da Vinci\textsuperscript{\textregistered} \cite{patsias2014feasibility,giataganas2015force}, has also been investigated. However, the combination of weak-depth perception and poor ergonomics, meand that it is still difficult for clinicians to perform a continuous and smooth microscopic scan during tele-operated surgery. Maintaining the probe in a specific orientation (normal to the tissue surface) and ensuring continuous contact with the tissue remains a challenging and tedious task. Therefore, the development of an autonomous scanning system for endomicroscopy would benefit its use in surgical operations by reducing the cognitive load on users and improving scanning accuracies over larger 3D surfaces.

In this work, we present a 6 degree-of-freedom (DoF) visual servoing method based on the da Vinci\textsuperscript{\textregistered} robot with the da Vinci Research Kit (dVRK). With our proposed approach, the robot is able to achieve smooth motion with sub-millimetre accuracy, and we have demonstrated that an autonomous large-area endomicroscopy scan can be performed by the robot over a user-defined area. In addition, to facilitate intraoperative tissue diagnosis and identification, a 3D visualisation method is proposed to fuse the 3D tissue surface with multiple microscopic mosaics on-the-fly, which is an improvement over the work of \cite{giataganas2014intraoperative}. This 3D fusion approach is designed to provide the surgeon with an intuitive and effective real-time visualisation of multi-scale imaging information, which supports surgical diagnosis and planning. Our framework has been tested on phantom and \emph{ex vivo} tissue experiments and the results demonstrate its potential clinical value. 

\begin{figure}[tb]
\vspace{0.2cm}
	\centering
	\includegraphics[width=\linewidth]{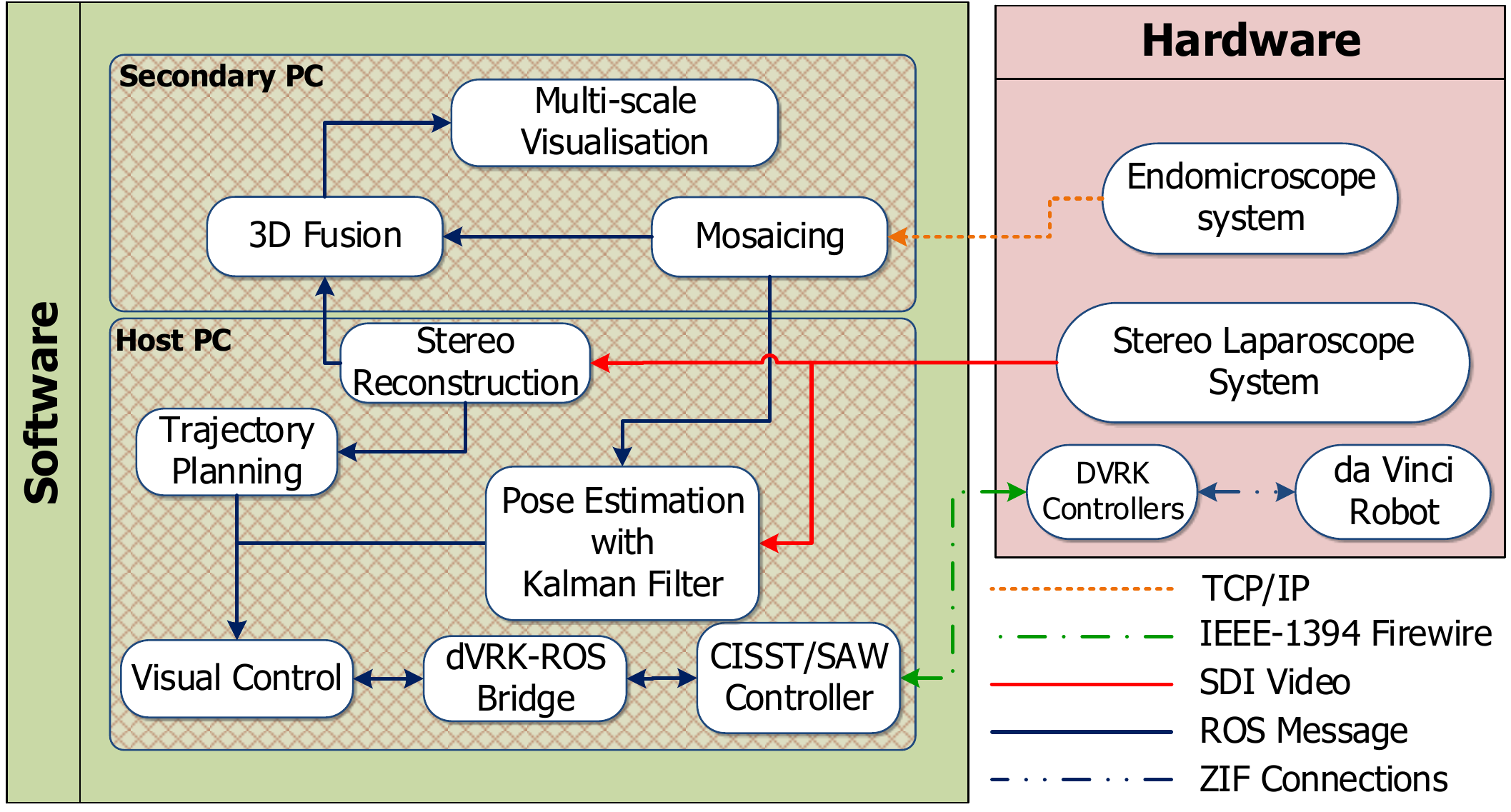}
	\caption{An overview of our proposed system framework for autonomous endomicroscopic scanning and 3D mosaicing.}
	\label{fig:system_framework}
\end{figure}

\section{Methodology} \label{Methodology}
\subsection{System Overview}
The hardware components of the proposed framework, highlighted in red in Fig. \ref{fig:system_framework}, consist of a patient side manipulator (PSM) of a da Vinci\textsuperscript{\textregistered} robot with dVRK controllers, a stereo laparoscope, and a custom endomicroscopy system. The dVRK controllers are connected to a host PC via a IEEE 1394 firewire interface in a daisy chain topology. The stereoscopic system provides SD (720x576) video streaming for both left and right channels at 25 Hz. In the host PC, the stereo stream is captured by a Kona 4 PCIe frame grabber (AJA Video System). 

The endomicroscope system is an in-house laser line-scanning fluorescence system, described fully in \cite{hughes2016line}, coupled to a Cellvizio UHD Probe (Mauna Kea Technologies). The probe, which consists of a 30,000 core fibre imaging bundle and a micro-lens, provides a FoV of 240 $\mu$m and a fibre-sampling limited resolution of approximately 2.4 $\mu$m. The line-scanning system provides optical sectioning, meaning that the endomicroscope collects light only from a in-focus plane approximately 20 $\mu$m in depth. Processed microscopic images are scaled to 300x300 pixels and served to the controller at 80 Hz via a TCP/IP connection. In addition to the host PC, a second PC with modest configuration is dedicated to 2D mosaicing and 3D visualisation.

The software component of the proposed framework runs on the Robot Operating System (ROS), highlighted in green in Fig. \ref{fig:system_framework}. The stereo images captured from the camera are used for 3D tissue surface reconstruction via a stereo matching method. The reconstructed 3D surface is used for two purposes: (1) to plan a scanning trajectory based on its positional and normal information; (2) to provide 3D fusion with the 2D image mosaics. The video stream from the laparoscope system is also used for estimating the pose of the endomicroscopy probe.

\begin{figure}[t]
	\vspace{0.2cm}
	\centering
	\includegraphics[width=\linewidth]{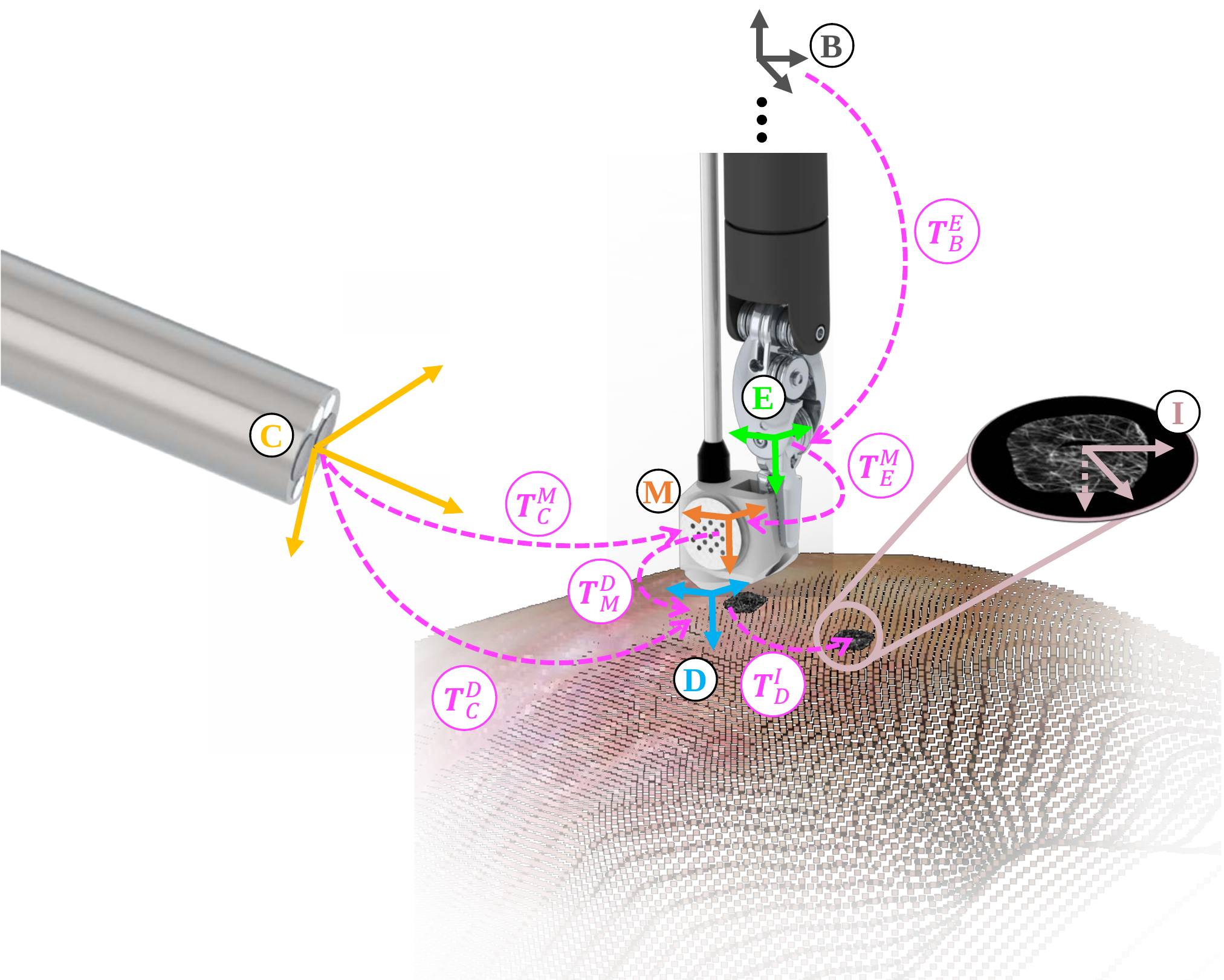}
	\caption{A description of coordinate systems (frames) and related transformations. $B$: robot base frame; $E$: end-effector frame; $M$: marker frame; $D$: probe tip frame; $C$: camera frame; $I$: mosaic point cloud frame.}
	\label{fig:frames}
\end{figure}

As shown in Fig. \ref{fig:frames}, the end-effector of the robot grasps an adapter that holds the endomicroscopy probe. A marker (KeyDot\textsuperscript{\textregistered}, Key Surgical, Minnesota) is attached to the adapter for pose estimation of the probe. The stereo system is calibrated using the method presented in \cite{Zhang2000} to obtain the intrinsic and extrinsic camera parameters such that the 3D depths of the tissue can be recovered via stereo matching. The trajectory planning consists of two parts: a global plan to guide the probe to the desired location of the scan (identified by the user), and a local plan to generate a mosaic. 

A visual control component closes the loop by comparing the current and desired probe poses and commanding the robot to minimise their difference, causing the probe to follow the planned trajectory. The robot's end-effector Cartesian pose is read and set via a dVRK-ROS component which is connected to a low level PID controller implemented by the SAW package using the cisst library \cite{kazanzidesf2014dvrk}.
Microscopic images captured from the endomicroscopy system are stitched together in real-time, using pairwise registration of images frames by template matching via normalised cross-correlation. A position estimate obtained from the mosaicking registration process is fused with the pose estimation from the camera using a Kalman filter, to provide probe position information used for closed loop control during the mosaic scanning. The image mosaic is also fused with the reconstructed surface on-the-fly to provide both macro- and micro-views of the scanned region. By providing accurate and robust pose estimation, this visual servoing approach permits smooth and accurate probe scanning, allowing the formation of contiguous image mosaics.

\subsection{Trajectory Planning Based on 3D Reconstruction}
\begin{figure}[t]
	\vspace{0.2cm}
	\centering
	\includegraphics[width=\linewidth]{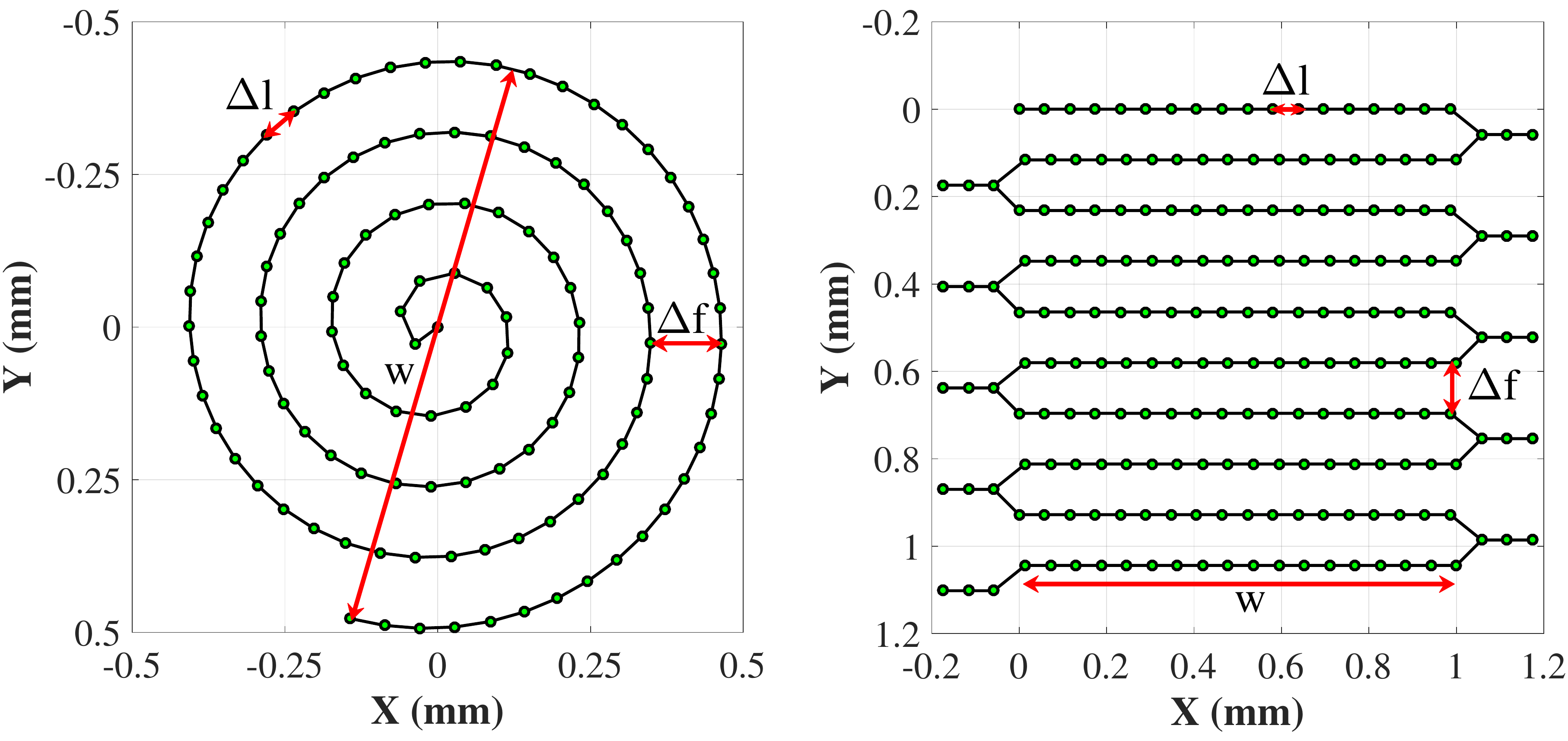}
	\caption{Illustration of two types of trajectory pattern for pCLE scanning: spiral (left) and raster (right). Three parameters that define the trajectory are labelled with red double-arrow line.}
	\label{fig:trajectory_plan}
\end{figure}

In this work, the scanning trajectory is planned to adapt to the 3D profile of the tissue surface. The Efficient LArge-scale Stereo (ELAS) \cite{geiger2011elas} approach has been adopted for 3D surface reconstruction. To determine the normal to a point on the surface, the problem is approximated by the problem of estimating the normal of a plane tangent to the surface, similar to \cite{RusuPhD}.

The scanning trajectory, which is a sequence of probe tip poses in the camera frame ($\mathbf{T}_{C}^{D}$), is planned in two stages: the global trajectory planning, which moves the probe to the starting pose for scanning, and the local trajectory planning, which performs high resolution scanning from the starting pose. To generate the probe tip poses in the global trajectory, we interpolate between the current probe tip pose at a standby position, and the desired starting pose for scanning. More specifically, linear interpolation and spherical linear interpolation are used for translation and rotation, respectively. The desired starting pose ensures that the endomicroscope is perpendicular to the tissue surface (using the surface normal from the 3D reconstruction) which is important for high quality microscopic image retrieval.

For the local scanning trajectory, we can assume that scanning surface is locally planar, which is reasonable as each individual scanning region is usually small (about 2x2mm) and most tissues are compliant. The local scanning trajectory can be planned on a 2D plane, with the aim of generating a contiguous mosaic over the desired area of tissue. We have investigated two types of scanning patterns, namely the raster and spiral trajectories. Example patterns are shown in Fig.\ref{fig:trajectory_plan}. A set of common parameters are used to specify the trajectory generation for both patterns. The diameter of the scanned region is defined by $ w $ which is typically 1-2 mm in our experiment. In order to reduce the likelihood of gaps in the mosaic image, but without excessive over-sampling, the overlap parameter $ \Delta f $ is set to half of the FoV of the endomicroscope, which is 240 $\mu$m. We also define a point spacing parameter $ \Delta l $ to set the interval between points in the trajectory.

For the raster pattern, each trajectory point $ (x_k, y_k) $ is defined as:
\begin{equation}
\label{eq:raster_traj}
\begin{split}
x_k &= \begin{cases}
k \cdot \Delta l, & k\in\left[ 0,\frac{w}{\Delta l}\right], \text{$ k $ in odd row}.\\
w-(k \cdot \Delta l), & k\in\left[ 0,\frac{w}{\Delta l}\right], \text{$ k $ in even row}.
\end{cases}
\\
y_k &= k \cdot \Delta f,  \hspace{14.6mm}k\in\big[  0,\hspace{-0.3mm}{\tfrac{w}{\Delta f}}\hspace{-0.3mm}\big]
\end{split}
\end{equation}
As the endomicroscopy probe maintains direct contact with the tissue at all times, minor tissue deformation would occur. Abrupt changes of direction (particularly moving between lines in the raster pattern) therefore results in loading and unloading of the tissue \cite{Erden2013}. In order to minimise this tissue deformation, we add an extra loading-unloading phase between two scanning lines, where the probe advances further than the limit of the planned trajectory, as shown in the Fig. \ref{fig:trajectory_plan}. This also helps compensating for the backlash of the robot which occurs during large directional changes. 

For the spiral pattern, we define:
\begin{equation} \label{eq:spiral_param}
\begin{cases}
b=\frac{\Delta f}{2\pi},\\
l_{sp}=\frac{b}{2} \cdot (\frac{w}{2b})^2.
\end{cases}
\end{equation}
Here, $ \Delta f $ is the distance between two successive spiral loops and $ l_{sp} $ is the total length of the spiral trajectory.

The k-th trajectory point in the spiral trajectory is given by:
\begin{equation} \label{eq:spiral_traj}
\begin{cases}
x_k=b \cdot \theta_k \cdot \cos \theta_k\\
y_k=b \cdot \theta_k \cdot \sin \theta_k\\
\end{cases}
\end{equation}
where $\theta_k$ is calculated using equation \ref{eq:spiral_param}:
\begin{equation} \label{eq:3}
\theta_k=\sqrt{\frac{2k\cdot \Delta l}{b}}, k\in\left[ 0,\frac{l_{sp}}{\Delta l}\right].
\end{equation}

\subsection{Pose Estimation with Kalman Filter}
Two types of imaging information are used for probe pose estimation: the laparoscopic camera image and the endomicroscope image. For probe pose estimation using the camera, a marker with asymmetric dots is attached to the adapter (see Fig. \ref{fig:frames}). In this work, we have used a vision-based detection-tracking method for marker recognition. The detection component is based on blob detection similar to \cite{pratt2012keydot}, which is applied on every image in the sequence. In order to improve the marker recognition rate, we have included a tracking component based on pyramidal optical flow \cite{bouguet2001opflow} to track the dots of the marker in time. Once the marker pattern is recognised in an image, the efficient perspective-n-points algorithm \cite{lepetit2009epnp} is applied to calculate the marker pose based on 2D-3D correspondences. The probe pose can be calculated as:
\begin{equation}
\mathbf{T}_{C}^{D} = \mathbf{T}_{C}^{M} \cdot \mathbf{T}_{M}^{D},
\end{equation}
where $ \mathbf{T}_{M}^{D} $ is a fixed transformation measured beforehand.

When using endomicroscope images for positional estimation of the probe, only translational motion on the image plane is considered. In order to filter the noisy camera measurements and fuse information from both imaging modalities, we applied a Kalman filter to obtain a final estimate of the probe probe.
Consider a discrete-time linear stationary signal model:
\begin{equation}
	\mathbf{x}(k)=\mathbf{A}\mathbf{x}(k-1)+\mathbf{w}(k)
\end{equation}
where $ \mathbf{x}(k) \in \mathbb{R}^{n} $ is the state vector. $ \mathbf{w}(k) \in \mathbb{R}^{n} $ is a sequence of process noise which is assumed to be drawn from a zero mean multivariate normal distribution whose covariance matrix is denoted by $ \mathbf{Q}(k) $. In this work, the state vector $ \mathbf{x}(k)=[t_x, t_y, t_z, \dot{t_x}, \dot{t_y}, \dot{t_z}, \alpha, \beta, \gamma, \dot{\alpha}, \dot{\beta}, \dot{\gamma}]^\intercal$ represents the position and orientation (in Euler angle form) of the probe together with their first derivatives (translational and angular velocity). The state transition matrix is defined as: $ \mathbf{A} = \left( \begin{smallmatrix} \mathbf{A}_T & \mathbf{0}\\ \mathbf{0}&\mathbf{A}_T \end{smallmatrix} \right)$ in which $ \mathbf{A}_T = \left( \begin{smallmatrix} \mathbf{I}_3 & \Delta t \cdot \mathbf{I}_3 \\ \mathbf{0}&\mathbf{I}_3 \end{smallmatrix} \right) $. The time interval between samples is denoted as $ \Delta t $.

The relationship between the measurements from sensors and the states is defined as:
\begin{equation}
	\mathbf{y}(k)=\mathbf{C}\mathbf{x}(k)+\mathbf{v}(k)
\end{equation}
where $ \mathbf{y}(k) \in \mathbb{R}^{l} $ is the measurement vector, and $ \mathbf{v}(k) \in \mathbb{R}^{l} $ is a sequence of observation noise which is assumed to be zero mean Gaussian white noise with covariance $  \mathbf{R}(k) $. Since we have two sensors, the measurement matrix $ C $ is defined separately for camera and pCLE:
\begin{equation}
\begin{cases}
\mathbf{C}_{c} =\left( \begin{smallmatrix} \mathbf{I}_3 & \mathbf{0} & \mathbf{0} & \mathbf{0}\\ \mathbf{0}& \mathbf{0} & \mathbf{I}_3 & \mathbf{0} \end{smallmatrix} \right),\\
\mathbf{C}_{p} = \left( \begin{smallmatrix} \mathbf{I}_2 & \mathbf{0} & \mathbf{0} & \mathbf{0} & \mathbf{0} & \mathbf{0} \end{smallmatrix} \right).\\
\end{cases}
\end{equation}

The error covariance of states is updated in the prediction stage as:
\begin{equation}
	\mathbf{P}(k|k-1) = \mathbf{A}\mathbf{P}(k-1|k-1)\mathbf{A}^{\intercal}+\mathbf{Q}(k-1).
\end{equation}
 
The Kalman gain for the camera and endomicroscope is given by:
\begin{equation}
\begin{cases}
\mathbf{K}_{c}(k) = \mathbf{P}(k|k-1)\mathbf{C}_{c}^{\intercal}\inv{(\mathbf{C}_{c}\mathbf{P}(k|k-1)\mathbf{C}_{c}^{\intercal}+\mathbf{R}_{c}(k))},\\
\mathbf{K}_{p}(k) = \mathbf{P}(k|k-1)\mathbf{C}_{p}^{\intercal}\inv{(\mathbf{C}_{p}\mathbf{P}(k|k-1)\mathbf{C}_{p}^{\intercal}+\mathbf{R}_{p}(k))},\\
\end{cases}
\end{equation}
The measurement residual is given by: $ \mathbf{r}(k) = \mathbf{y}(k)-\mathbf{C} \mathbf{x}(k-1)$. The correction is then performed using:
\begin{equation}
\begin{cases}
\mathbf{x}(k) = \mathbf{x}(k|k-1)+\frac{1}{2}(\mathbf{K}_c(k)\mathbf{r}_c(k) + \mathbf{K}_p(k)\mathbf{r}_p(k)),\\
\mathbf{P}(k) = (\mathbf{I} - \frac{1}{2}(\mathbf{K}_c(k) \mathbf{C}_c + \mathbf{K}_p(k) \mathbf{C}_p)) \mathbf{P}(k|k-1).\\
\end{cases}
\end{equation}

During the execution of global trajectory to the scanning region, there is no endomicroscopy image information (the probe is not in contact with the tissue) and thus only the camera data is fed into the Kalman filter for that case.

\subsection{Closed Loop Scanning}
In our framework, we compute the hand-tool transformation $ \mathbf{T}_{E}^{M} $ between the end-effector and marker on the adaptor via a standard hand-eye calibration approach \cite{tsai1989handeye}. The advantage of using the hand-tool transformation is that it will not be affected when the robot and camera are repositioned, making it more practical during surgical procedures. 
Although $ \mathbf{T}_{E}^{M} $ would change slightly if the adapter is released and re-grasped, we found that the resulting small error (a few millimetres) in the hand-tool transformation is acceptable for the visual servoing control.

Given a current estimated probe pose $\mathbf{T}_{C}^{D}$ and a desired pose $\mathbf{T}_{C}^{D^{\star}}$ in the camera frame (in the planned trajectory), a new command pose can be calculated for the robot to minimise the difference.
The relative probe pose, denoted as \(\mathbf{T}_{D}^{D^{\star}}\), is given by:
\begin{equation}
\mathbf{T}_{D}^{D^{\star}} = \inv{\left(\mathbf{T}_{C}^{D} \right)} \cdot \mathbf{T}_{C}^{D^{\star}}
\end{equation}
where $ \mathbf{T}_{C}^{D} $ and $ \mathbf{T}_{C}^{D^{\star}} $ are obtained from the Kalman filter and the planned trajectory, respectively. Since the local trajectory is planned on the mosaic image coordinate, we need transformation $  \mathbf{T}_{D}^{I} = \left( \begin{smallmatrix} \mathbf{R}_{D}^{I} & \mathbf{0}\\ 0 & 1 \end{smallmatrix} \right)$ to convert the trajectory from 2D to 3D. $\mathbf{R}_{D}^{I}$ is essentially a 2D rotation between the probe and the mosaic image, which can be calibrated by driving the robot with a horizontal line scan.

With this, we can calculate the robot command $ \mathbf{T}_{B}^{E} $ that minimises the relative probe pose:
\begin{equation}
\mathbf{T}_{B}^{E^{\star}} = \mathbf{T}_{B}^{E} \cdot \mathbf{T}_{E}^{M} \cdot \mathbf{T}_{M}^{D} \cdot \mathbf{T}_{D}^{D^{\star}} \cdot \inv{\left(\mathbf{T}_{M^{\star}}^{D^{\star}}\right)} \cdot \inv{\left(\mathbf{T}_{E^{\star}}^{M^{\star}}\right)}
\end{equation}

The vision-based correction is performed on every pose along the planned trajectory. To produce a smooth scanning motion, the trajectory between two adjacent poses is interpolated and executed using the robot kinematic model.

\subsection{Endoscopic 3D Fusion and Visualisation}
Using the 3D reconstruction of the scanning region and the probe pose in camera coordinates, we are able to register the mosaic images with the tissue surface. This enables us to create a multi-scale map that links macro- and micro-views, which are obtained from the stereo camera and the endomicroscope, respectively. For endomicroscopy image mosaicing, we have used an approach similar to standard real-time techniques discussed in \cite{hughes2015high}, using normalised cross-correlation to estimate the relative shift between each pair of consecutive frames. To avoid overwriting information from previous frames, and minimise boundary effects where images overlap, blending is applied. The mosaic image is then converted to a 3D mosaic point cloud in which each point represents a pixel in the mosaic image.

In order to register the mosaic point cloud with the surface, we need to find $ \mathbf{T}_{I}^{C} $. As illustrated in \ref{fig:frames} it can be calculated as: 
\begin{equation}
\mathbf{T}_{C}^{I} = \mathbf{T}_{C}^{M_0} \cdot \mathbf{T}_{M_0}^{D} \cdot  \mathbf{T}_{D}^{I},
\end{equation}
where $ M_0 $ represents the coordinate frame of the marker at the beginning of the local scan.
With $ \mathbf{T}_{I}^{C} $ for each scanning region, we are able to register mosaics from multiple region scans onto the same 3D surface. The 3D mosaic point cloud is updated on-the-fly when the microscopic images are mosaicked. An illustration of this multi-scale 3D fusion can be seen in Fig. \ref{fig:fusion}.

In this paper, we provide three ways to visualise the scanning and mosaicing on-the-fly: (1) An augmented view from the camera that allows showing both the robot and target tissue by overlaying the endomicroscope images onto the tracking marker on the adaptor, as shown in Fig. \ref{fig:phantomexp}(b) and (d); (2) A real-time, continuously updated 2D mosaic map; (3) An online  3D fusion of mosaics onto the reconstructed surface. These methods provide a clear and intuitive visualisation for the surgeon, aimed at assisting surgical planning.

\section{Experiments and Results} \label{Experiment}

\begin{table}[t]
\vspace{0.2cm}
	\centering
	\caption{Stereo Reconstruction Accuracy}
	\label{Table_stereo}
	\begin{tabular}{ccccc}
		\hline
		&     & \textbf{ELAS} \cite{geiger2011elas} & \textbf{COCV}  \cite{chang2013stereo} &  \\ \hline
		\multirow{2}{*}{Cardiac1} & MAE (mm) & 0.89 $\pm$ 0.70	& 1.24$\pm$ 0.89 &  \\
											 & RMS (mm) & 1.31  $\pm$ 0.98	& 1.85 $\pm$ 0.82 &  \\ \hline
		\multirow{2}{*}{Cardiac2} & MAE (mm) & 1.21  $\pm$ 1.56	& 1.47 $\pm$ 1.23 &  \\
											 & RMS (mm) & 1.77 $\pm$ 2.16	&2.66 $\pm$ 1.47 &  \\ \hline
	\end{tabular}
\end{table}

 \begin{table}[t]
 	\vspace{0.2cm}
 	\centering
 	\caption{Camera Visual Servoing Error (Mean $\pm$ Std)}
 	{\scriptsize
 		\begin{tabular}{ccccccccc}
 			\hline\\ [-0.2cm]
 			& \multicolumn{2}{c}{\textbf{Translation Error (mm)}} & \multicolumn{2}{c}{\textbf{Rotation Error (degrees)}}\\[0.1cm]
 			\textbf{Trial} & \textbf{VS} & \textbf{KM} & \textbf{VS} & \textbf{KM}\\ 
 			\hline\\[-0.2cm]
 			
 			\textbf{Trial 1} & 0.238 $\pm$ 0.06 & 0.863 $\pm$ 0.26 & 0.89 $\pm$ 0.55 & 4.51 $\pm$ 1.45 \\
 			\textbf{Trial 2} & 0.222 $\pm$ 0.08 & 0.360 $\pm$ 0.21 & 1.21 $\pm$ 0.53 & 3.57 $\pm$ 0.79 \\
 			\textbf{Trial 3} & 0.189 $\pm$ 0.08 & 0.507 $\pm$ 0.16 & 1.19 $\pm$ 0.55 & 1.96 $\pm$ 1.00 \\
 			\textbf{Trial 4} & 0.185 $\pm$ 0.07 & 0.588 $\pm$ 0.37 & 1.54 $\pm$ 0.46 & 5.54 $\pm$ 1.27 \\
 			\textbf{Trial 5} & 0.172 $\pm$ 0.06 & 0.462 $\pm$ 0.27 & 1.44 $\pm$ 0.55 & 5.62 $\pm$ 1.11 \\
 			\textbf{Trial 6} & 0.226 $\pm$ 0.07 & 0.915 $\pm$ 0.40 & 1.57 $\pm$ 0.49 & 5.22 $\pm$ 1.08 \\
 			\textbf{Trial 7} & 0.247 $\pm$ 0.06 & 0.909 $\pm$ 0.37 & 1.11 $\pm$ 0.48 & 4.05 $\pm$ 0.87 \\
 			\textbf{Trial 8} & 0.303 $\pm$ 0.09 & 0.542 $\pm$ 0.17 & 1.25 $\pm$ 0.50 & 0.79 $\pm$ 0.52 \\
 			\textbf{Trial 9} & 0.165 $\pm$ 0.10 & 0.484 $\pm$ 0.13 & 0.84 $\pm$ 0.45 & 3.39 $\pm$ 0.80 \\
 			\textbf{Trial 10} & 0.164 $\pm$ 0.07 & 0.666 $\pm$ 0.13 & 1.21 $\pm$ 0.57 & 1.44 $\pm$ 1.04 \\
 			\hline
 			\textbf{Total}  & \textbf{0.211} $\pm$ \textbf{0.07} & \textbf{0.630} $\pm$ \textbf{0.25} & \textbf{1.226} $\pm$ \textbf{0.51} & \textbf{3.61} $\pm$ \textbf{0.99} \\
 			\hline
 		\end{tabular}}
 		\label{Table_Global_Error}
 	\end{table}

\subsection{Stereo Validation}
For assessing the accuracy of ELAS stereo reconstruction, we have provided a comparison of the method with the state-of-the-art \cite{chang2013stereo} on the Hamlyn cardiac datasets 
(\url{http://hamlyn.doc.ic.ac.uk/vision/}). 
The results are provided in Table \ref{Table_stereo}, and show that the ELAS method provides competitive accuracies.

\subsection{Hand-eye calibration accuracy}
Due to the kinematic error of a tendon driven arm such as the da Vinci robot, the hand-eye calibration would only be accurate within a space where the calibration data has been taken. Given the calibrated hand-tool transformation $\mathbf{T}_{E}^{M}$ and an initial marker pose $ \mathbf{T}_{C}^{M_{0}} $, we can predict a marker pose using robot forward kinematics:

\begin{equation}\label{eq:handeye_error}
\mathbf{\hat{T}}_{C}^{M} = \mathbf{T}_{C}^{M_{0}} \cdot \mathbf{T}_{M_{0}}^{E_{0}} \cdot \mathbf{T}_{E_{0}}^{B} \cdot \mathbf{T}_{B}^{E} \cdot \mathbf{T}_{E}^{M}.
\end{equation}

The ground truth of the marker pose $ \mathbf{T}_{C}^{M} $ was obtained from manual annotation.
We calculate the error between the ground truth and the pose estimated by the hand-eye model as described by Eq. \ref{eq:handeye_error}. The error pose is computed as: $ \Delta\mathbf{T}_{\hat{M}}^{M} = \inv{\left( \mathbf{\hat{T}}_{C}^{M} \right)} \cdot \mathbf{T}_{C}^{M} $. Starting from the initial pose $ \mathbf{T}_{C}^{M_{0}} $, as the marker moves within the workspace for the scanning task, we calculate two values: translational error and rotational error. The translational error is the Euclidean norm of the translational part of $ \Delta\mathbf{T}_{\hat{M}}^{M} $ while the rotational error is the mean of the Euler angle of the rotational part. As presented in \ref{tab:handeye_error}, the translational and rotational errors are approximately 1.5 mm and 0.7 degrees respectively, which is sufficient to support sub-millimetre-scale accuracy for the visual servoing.

\begin{table}[]
	\centering
	\caption{Error of Hand-eye calibration}
	\label{table_handeye_error}
	\begin{tabular}{@{}ccc@{}}
		\toprule
		& Translation (mm) & Rotation (deg)        \\ \midrule
		MAE  & 1.33 $\pm$ 0.78 & 0.64 $\pm$ 0.37\\
		RMSE & 1.54  & 0.74\\ \bottomrule
	\end{tabular}
	\label{tab:handeye_error}
\end{table}

\subsection{Accuracy of Visual Servoing}

To quantitatively validate the camera visual servoing approach, we used the Optotrak Certus system (Northern Digital Inc, Canada) which achieves 0.1 mm accuracy. Optical sensors were attached to a custom probe adapter, so that a rigid transformation could be found between the optical sensors and the marker frame. In this study, we ran ten trials for validation. Each trial consisted of two runs that included trajectory-following with Visual Servoing (VS) and Kinematic Motion (KM) only. The planned trajectories, which are defined in the camera coordinates, were used as the ground truth, and both the optical tracked results of two runs were transformed into the camera coordinate for comparison. To this end, we assume the starting pose of two runs should align with the first pose in the ground truth trajectory. For each pair of measured and ground truth poses at the same time instance, we calculate the pose difference as the error. For translational errors, we compute the Euclidean distance, and for rotational errors, we use the average of the Euler angles from the x, y and z axes.
The measured translational errors and rotational errors are summarised in Table \ref{Table_Global_Error}. 
Both translational and rotational accuracies using the camera visual servoing are improved over kinematic-only control. The motion using the camera visual servoing was able to achieve on average approximately 0.21 mm for the translational error and 1.23 degrees for the rotational error.

\begin{figure}[tb]
	\centering
	\includegraphics[width=\linewidth]{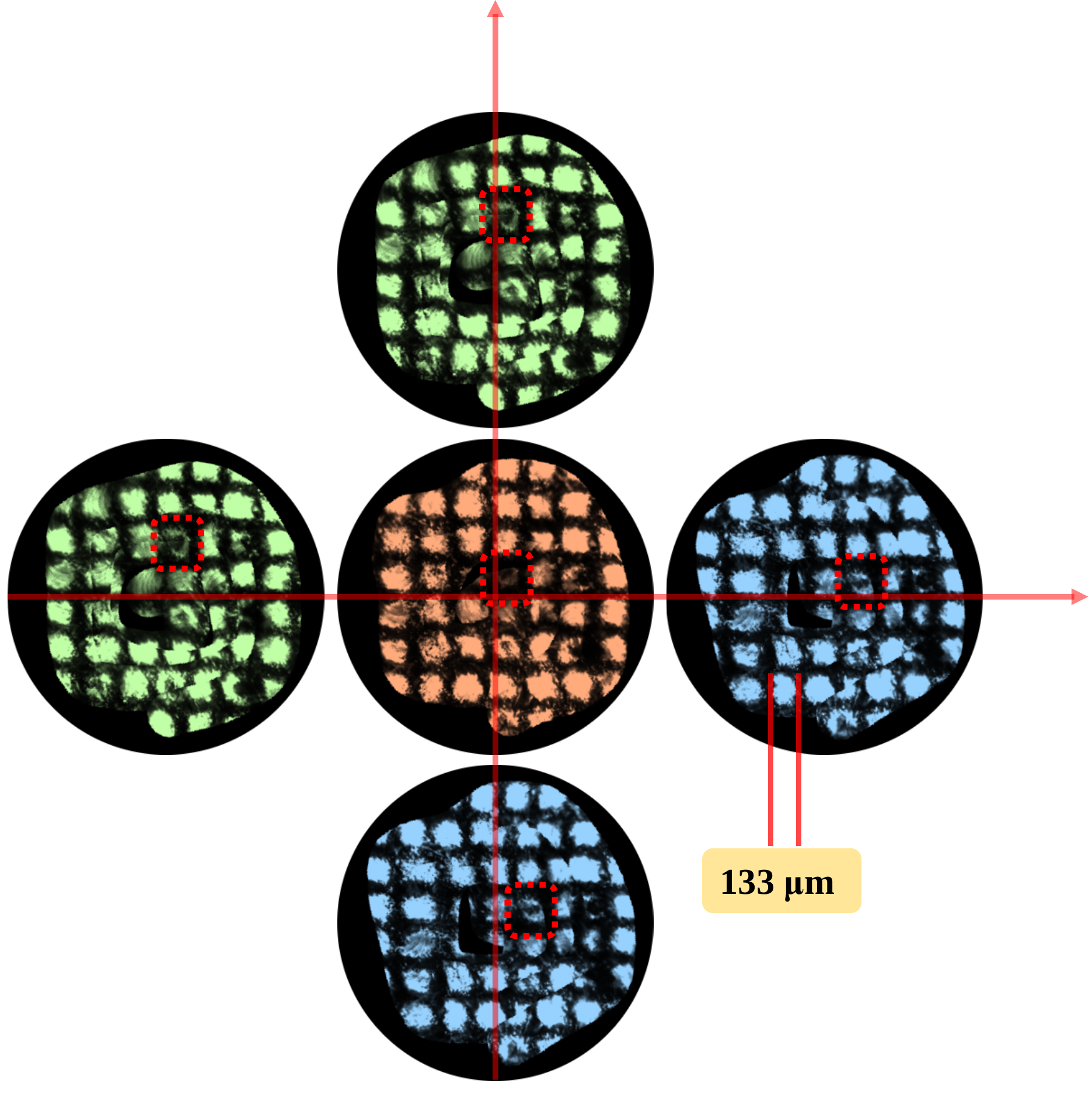}
	\caption{Demonstration of mosaic results on a printed grid phantom with black holes (outlined by the dashed rectangle). Three trials are identified by three different colours and a red guide axis is placed for assisting comparison.\vspace{-0.2cm}}
	\label{fig:global_mosaic_comp}
\end{figure}

Since the camera-based visual servoing is used to move the probe to an initial position for scanning, we can qualitatively evaluate the repeatability by comparing the differences among multiple mosaics acquired at the same nominal position. We ran three trials, in which the probe was moved to the same position before scanning and mosaicing over a printed grid phantom. As shown in Fig. \ref{fig:global_mosaic_comp}, the mosaic images from different trials (identified by the colour) are well-aligned at their centres, both horizontally and vertically, and the camera visual servoing provides good repeatability with only about 50 to 150 $\mu$m deviation. This shows good correspondence with the quantitative accuracy estimated above (0.211 mm).

The mosaic images in the Fig. \ref{fig:global_mosaic_comp}, we also provide us with the accuracy of camera and endomicroscope combined servoing. This is performed by comparing the original layout of the grid pattern with the obtained mosaicked grid, and in our experiment, the error is around 60 $\mu$m including the uncertainty introduced by the laser printer.

\begin{figure}[t]
      \centering
      \includegraphics[width=\linewidth]{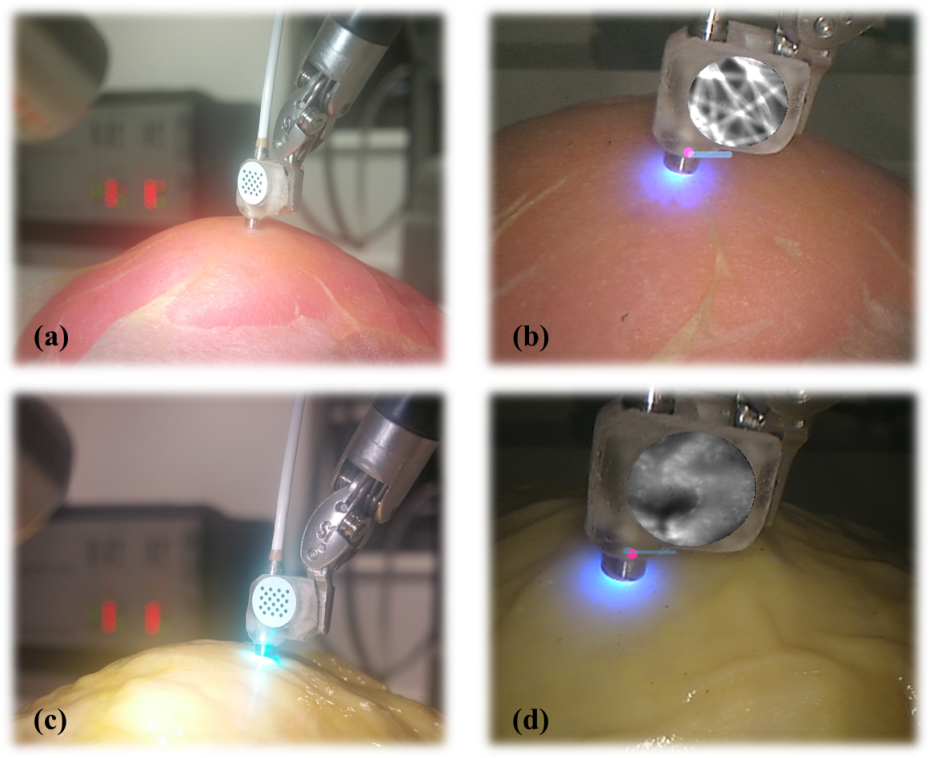}
      \caption{Snapshots of the phantom and \emph{ex vivo} experiments. (a) and (c) Experimental setups; (b) and (d) A snapshot of the surgical view captured by the laparoscope. The KeyDot\textsuperscript{\textregistered} marker was overlaid with the streaming microsopic images for augmented reality. \vspace{-0.2cm}}
      \label{fig:phantomexp}
\end{figure}
   
 \begin{figure}[t]
 \vspace{0.2cm}
	\centering
	\includegraphics[width=\linewidth]{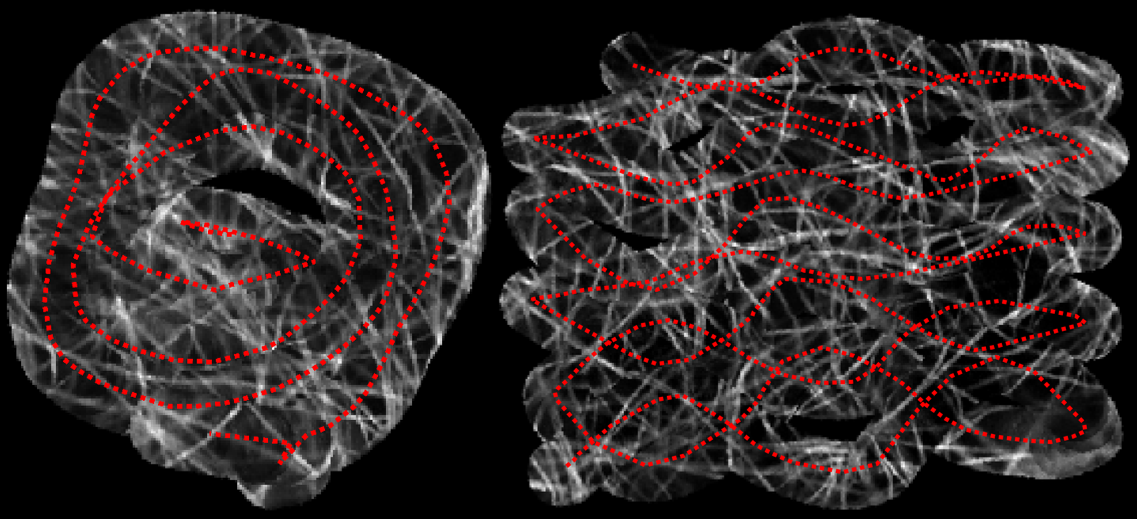}
	\caption{Comparison of spiral and raster trajectories. The actual trajectories measured on the mosaic images are shown in red dotted line.}
	\label{fig:spiral_raster_comp}
\end{figure}

\begin{figure*}[t]
      \centering
      \includegraphics[width=\linewidth]{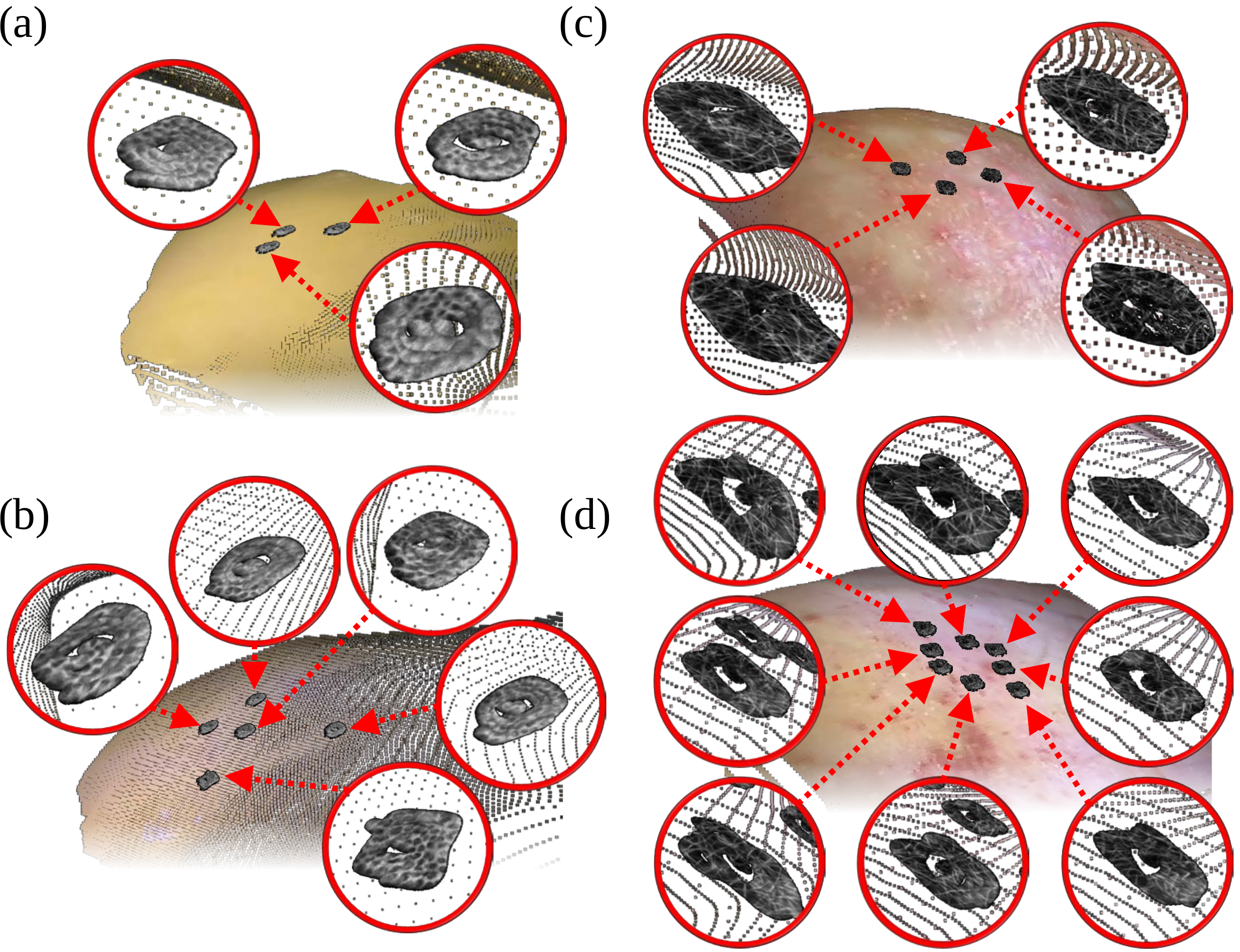}
      \caption{Exemplar 3D fusion results. (a) and (b) Online multi-region fusion results from the \emph{ex vivo} gastric mucosa experiment; (c) and (d) Online multi-region fusion results from phantom experiment.}
      \label{fig:fusion}
\end{figure*}
\subsection{Phantom and Ex Vivo Experiment}
The experiment steps can be summarised as follow:
\begin{enumerate}
	\item A stereo reconstruction of current surgical scene is generated.
	\item A user places the probe at a standby location.
	\item A user selects a target scanning region with the global and local trajectories being generated accordingly.
	\item The robot moves the probe along the global trajectory with servoing from the laparoscopic camera images.
	\item Probe arrives at the starting position for the mosaic scan over the desired region; robot drives probe to follow local trajectory using both microscopic and laparoscopic images for servoing.
	\item After finishing the scan, the probe returns to the standby location, ready to scan the next region.
\end{enumerate}

To validate the approach, we have performed autonomous scans on a custom-made PVA cryogel kidney phantom. In order to produce high quality microscopic images, lens tissue paper stained with the topical fluorescent contrast agent acriflavine, was placed on top of the surface. A snapshot of the experiment is provided in Fig. \ref{fig:phantomexp}(a-b). Scanning tasks were then performed on the phantom surface using both spiral and raster trajectories. An example of the 2D mosaics obtained is shown in Fig. \ref{fig:spiral_raster_comp}. The actual trajectory line drawn on the mosaic image shows that the robot suffers from the raster motion due to backlash, as the raster pattern consists of abrupt changes of motion direction. The spiral pattern is considered to be superior as the motion direction changes gradually.

The 3D fusion results are presented in Fig. \ref{fig:fusion} which demonstrates that our approach is able to perform multi-region scanning. These results have shown our proposed framework is able to provide continuous image mosaicing, benefiting from the smooth and accurate motion generated by the visual servoing approach. 
Experiments were also performed using \emph{ex vivo} animal tissue (see Fig. \ref{fig:phantomexp}(c-d)). The gastric mucosa of a porcine stomach tissue sample was again stained with the topical acriflavine contrast agent. Examples of the 3D fusion results are provided in Fig. \ref{fig:fusion}. Both the phantom and \emph{ex vivo} experiments have demonstrated that our framework is capable of performing autonomous endomicroscopy scanning and generating high quality image mosaics. Again, due to the advantages of global surface reconstruction and visual servo control, we are able to scan multiple regions on the same surface. This also provides a global-local mapping where user can not only see each scanning location from macro-scale but also see a zoom-in mosaic image.

\subsection{Discussion}
Whilst one might expect that the accuracy of stereo reconstruction may be insufficient for an endomicroscopy scanning task, we found in the experiments that most of the scanning tasks can be completed successfully. However, it was observed during the stereo validation that the recovered surface tends to have a constant offset despite good shape matching to the ground truth. To address this erroneous offset, we adjusted the probe tip to marker transformation $\mathbf{T}_{D}^{M}$ to ensure continuous contact with the surface. This solution is not ideal and would cause problems when the reconstruction error becomes too large ($>2.0mm$). This is because the large error would cause the probe to either lose contact with the surface or apply too much pressure, leading to excessive deformation. In addition, we also observed that the image quality tends to decrease as the contact force increases. A comprehensive solution to this problem, would be to perform additional visual servoing based on the quality of microscopic images. With this, the image quality could then be used to detect loss of contact or excessive pressure, as well as to adjust the robot position accordingly.

The proposed framework presents good repeatability and accuracy, where all trials present consistently low errors, sufficient to allow endomicroscopy scanning tasks to be completed. The mosaicing results from both phantom and \emph{ex vivo} experiments, shown in Fig. \ref{fig:fusion}, have demonstrated that the system is capable of scanning multiple regions while obtaining good quality mosaicing results. The mosaics generated in this work are a significant improvement over the basic linear mosaics produced by manually operated endomicroscopes, suggesting improved clinical value for intraoperative pathological analysis and surgical planning.

\section{Conclusions}
In this paper, an autonomous framework using the da Vinci\textsuperscript{\textregistered} robot is proposed for endomicroscopic mosaicing and online fusion of 3D reconstructed macroscopic images with microscopic images. We have used both laparscopic camera and endomicroscopy images to close the loop to achieve continuous and smooth scanning. In addition, the proposed framework provides a comprehensive visualisation scheme for users, which includes: (1) an overlay of microscopic images into the surgical view for augmented reality; (2) an online updating 2D mosaic image; and (3) 3D fusion of the mosaic images with the tissue surface on-the-fly. These visualisation options have a range of potential benefits for intraoperative tissue analysis and surgical planning, which will be explored in future work. The proposed framework has been tested on phantoms and \emph{ex vivo} tissue, and quantitative results obtained from an external tracking system have shown that our framework can achieve accuracy better than the field-of-view of the endomicroscope, thus permitting smooth and accurate microscopic scanning.

\section*{ACKNOWLEDGMENT}

The authors would like to thank Simon DiMaio from Intuitive Surgical Inc for providing the CAD model of the large needle driver, Konrad Leibrandt for his assistance with the software, and Stamatia Giannarou for valuable discussions.

\bibliographystyle{IEEEtran}
\bibliography{IEEEabrv,main}

\begin{thebibliography}{10}
\providecommand{\url}[1]{#1}
\csname url@samestyle\endcsname
\providecommand{\newblock}{\relax}
\providecommand{\bibinfo}[2]{#2}
\providecommand{\BIBentrySTDinterwordspacing}{\spaceskip=0pt\relax}
\providecommand{\BIBentryALTinterwordstretchfactor}{4}
\providecommand{\BIBentryALTinterwordspacing}{\spaceskip=\fontdimen2\font plus
\BIBentryALTinterwordstretchfactor\fontdimen3\font minus
  \fontdimen4\font\relax}
\providecommand{\BIBforeignlanguage}[2]{{%
\expandafter\ifx\csname l@#1\endcsname\relax
\typeout{** WARNING: IEEEtran.bst: No hyphenation pattern has been}%
\typeout{** loaded for the language `#1'. Using the pattern for}%
\typeout{** the default language instead.}%
\else
\language=\csname l@#1\endcsname
\fi
#2}}
\providecommand{\BIBdecl}{\relax}
\BIBdecl

\bibitem{moustris2011autosurgreview}
G.~Moustris, S.~Hiridis, K.~Deliparaschos, and K.~Konstantinidis, ``Evolution
  of autonomous and semi-autonomous robotic surgical systems: a review of the
  literature,'' \emph{The International Journal of Medical Robotics and
  Computer Assisted Surgery}, vol.~7, no.~4, pp. 375--392, 2011.

\bibitem{billings2012us}
S.~Billings, N.~Deshmukh, H.~J. Kang, R.~Taylor, and E.~M. Boctor, ``System for
  robot-assisted real-time laparoscopic ultrasound elastography,'' in
  \emph{SPIE Medical Imaging}.\hskip 1em plus 0.5em minus 0.4em\relax
  International Society for Optics and Photonics, 2012, pp. 83\,161W--83\,161W.

\bibitem{ruszkowski2015dvrk}
A.~Ruszkowski, O.~Mohareri, S.~Lichtenstein, R.~Cook, and S.~Salcudean, ``On
  the feasibility of heart motion compensation on the davinci{\textregistered}
  surgical robot for coronary artery bypass surgery: Implementation and user
  studies,'' in \emph{Robotics and Automation (ICRA), 2015 IEEE International
  Conference on}.\hskip 1em plus 0.5em minus 0.4em\relax IEEE, 2015, pp.
  4432--4439.

\bibitem{padoy2011humansuture}
N.~Padoy and G.~D. Hager, ``Human-machine collaborative surgery using learned
  models,'' in \emph{Robotics and Automation (ICRA), 2011 IEEE International
  Conference on}.\hskip 1em plus 0.5em minus 0.4em\relax IEEE, 2011, pp.
  5285--5292.

\bibitem{Murali2015}
A.~Murali, S.~Sen, B.~Kehoe, A.~Garg, S.~McFarland, S.~Patil, W.~D. Boyd,
  S.~Lim, P.~Abbeel, and K.~Goldberg, ``Learning by observation for surgical
  subtasks: Multilateral cutting of 3d viscoelastic and 2d orthotropic tissue
  phantoms,'' in \emph{Robotics and Automation (ICRA), 2015 IEEE International
  Conference on}, May 2015, pp. 1202--1209.

\bibitem{pratt2015autonomous}
P.~Pratt, A.~Hughes-Hallett, L.~Zhang, N.~Patel, E.~Mayer, A.~Darzi, and G.-Z.
  Yang, ``Autonomous ultrasound-guided tissue dissection,'' in \emph{Medical
  Image Computing and Computer-Assisted Intervention--MICCAI 2015}.\hskip 1em
  plus 0.5em minus 0.4em\relax Springer, 2015, pp. 249--257.

\bibitem{hu2015semibrain}
D.~Hu, Y.~Gong, B.~Hannaford, and E.~J. Seibel, ``Semi-autonomous simulated
  brain tumor ablation with ravenii surgical robot using behavior tree,'' in
  \emph{Robotics and Automation (ICRA), 2015 IEEE International Conference
  on}.\hskip 1em plus 0.5em minus 0.4em\relax IEEE, 2015, pp. 3868--3875.

\bibitem{wallace2010preliminary}
M.~B. Wallace, P.~Sharma, C.~Lightdale, H.~Wolfsen, E.~Coron, A.~Buchner,
  M.~Bajbouj, A.~Bansal, A.~Rastogi, J.~Abrams \emph{et~al.}, ``Preliminary
  accuracy and interobserver agreement for the detection of intraepithelial
  neoplasia in barrett's esophagus with probe-based confocal laser
  endomicroscopy,'' \emph{Gastrointestinal endoscopy}, vol.~72, no.~1, pp.
  19--24, 2010.

\bibitem{meining2011pcle}
A.~Meining, Y.~K. Chen, D.~Pleskow, P.~Stevens, R.~J. Shah, R.~Chuttani,
  J.~Michalek, and A.~Slivka, ``Direct visualization of indeterminate
  pancreaticobiliary strictures with probe-based confocal laser endomicroscopy:
  a multicenter experience,'' \emph{Gastrointestinal endoscopy}, vol.~74,
  no.~5, pp. 961--968, 2011.

\bibitem{Ye2013}
M.~Ye, S.~Giannarou, N.~Patel, J.~Teare, and G.-Z. Yang, ``Pathological site
  retargeting under tissue deformation using geometrical association and
  tracking,'' in \emph{Medical Image Computing and Computer-Assisted
  Intervention--MICCAI 2013}.\hskip 1em plus 0.5em minus 0.4em\relax Springer,
  2013, pp. 67--74.

\bibitem{Latt2011}
W.~T. Latt, R.~Newton, M.~Visentini-Scarzanella, C.~Payne, D.~Noonan, J.~Shang,
  and G.-Z. Yang, ``A hand-held instrument to maintain steady tissue contact
  during probe-based confocal laser endomicroscopy,'' \emph{Biomedical
  Engineering, IEEE Transactions on}, vol.~58, no.~9, pp. 2694--2703, Sept
  2011.

\bibitem{rosa2013building}
B.~Rosa, M.~S. Erden, T.~Vercauteren, B.~Herman, J.~Szewczyk, and G.~Morel,
  ``Building large mosaics of confocal edomicroscopic images using visual
  servoing,'' \emph{Biomedical Engineering, IEEE Transactions on}, vol.~60,
  no.~4, pp. 1041--1049, 2013.

\bibitem{dwyer2015miniaturised}
G.~Dwyer, P.~Giataganas, P.~Pratt, M.~Hughes, and G.-Z. Yang, ``A miniaturised
  robotic probe for real-time intraoperative fusion of ultrasound and
  endomicroscopy,'' in \emph{Robotics and Automation (ICRA), 2015 IEEE
  International Conference on}.\hskip 1em plus 0.5em minus 0.4em\relax IEEE,
  2015, pp. 1196--1201.

\bibitem{Zuo2015breast}
S.~Zuo, M.~Hughes, C.~Seneci, T.~Chang, and G.-Z. Yang, ``Towards
  intraoperative breast endomicroscopy with a novel surface scanning device,''
  \emph{Biomedical Engineering, IEEE Transactions on}, vol.~PP, no.~99, pp.
  1--1, 2015.

\bibitem{patsias2014feasibility}
A.~Patsias, L.~Giraldez-Rodriguez, A.~Polydorides, R.~Richards-Kortum,
  S.~Anandasabapathy, T.~Quang, A.~Sikora, and B.~Miles, ``Feasibility of
  transoral robotic-assisted high resolution microendoscopic imaging of
  oropharyngeal squamous cell carcinoma,'' \emph{Head \& neck}, 2014.

\bibitem{giataganas2015force}
P.~Giataganas, M.~Hughes, and G.-Z. Yang, ``Force adaptive robotically assisted
  endomicroscopy for intraoperative tumour identification,''
  \emph{International journal of computer assisted radiology and surgery}, pp.
  1--8, 2015.

\bibitem{giataganas2014intraoperative}
P.~Giataganas, C.~Bergeles, P.~Pratt, M.~Hughes, A.~Darzi, and G.-Z. Yang,
  ``Intraoperative 3d fusion of microscopic and endoscopic images in transanal
  endoscopic microsurgery,'' in \emph{The Hamlyn Symposium on Medical
  Robotics}, 2014, p.~35.

\bibitem{hughes2016line}
M.~Hughes and G.-Z. Yang, ``Line-scanning fiber bundle endomicroscopy with a
  virtual detector slit,'' \emph{Biomedical Optics Express}, vol.~7, no.~6, pp.
  2257--2268, 2016.

\bibitem{Zhang2000}
Z.~Zhang, ``A flexible new technique for camera calibration,'' \emph{Pattern
  Analysis and Machine Intelligence, IEEE Transactions on}, vol.~22, no.~11,
  pp. 1330--1334, Nov 2000.

\bibitem{kazanzidesf2014dvrk}
P.~Kazanzides, Z.~Chen, A.~Deguet, G.~S. Fischer, R.~H. Taylor, and S.~P.
  DiMaio, ``An open-source research kit for the da vinci{\textregistered}
  surgical system,'' in \emph{Robotics and Automation (ICRA), 2014 IEEE
  International Conference on}.\hskip 1em plus 0.5em minus 0.4em\relax IEEE,
  2014, pp. 6434--6439.

\bibitem{geiger2011elas}
A.~Geiger, M.~Roser, and R.~Urtasun, ``Efficient large-scale stereo matching,''
  in \emph{Computer Vision--ACCV 2010}.\hskip 1em plus 0.5em minus 0.4em\relax
  Springer, 2011, pp. 25--38.

\bibitem{RusuPhD}
R.~B. Rusu, ``{Semantic 3D Object Maps for Everyday Manipulation in Human
  Living Environments},'' Ph.D. dissertation, Technische Universit\"at
  M\"unchen, 2009.

\bibitem{Erden2013}
M.~S. Erden, B.~Rosa, J.~Szewczyk, and G.~Morel, ``Understanding soft-tissue
  behavior for application to microlaparoscopic surface scan,'' \emph{IEEE
  Transactions on Biomedical Engineering}, vol.~60, no.~4, pp. 1059--1068,
  April 2013.

\bibitem{pratt2012keydot}
P.~Pratt, A.~Di~Marco, C.~Payne, A.~Darzi, and G.-Z. Yang, ``Intraoperative
  ultrasound guidance for transanal endoscopic microsurgery,'' in \emph{Medical
  Image Computing and Computer-Assisted Intervention--MICCAI 2012}.\hskip 1em
  plus 0.5em minus 0.4em\relax Springer, 2012, pp. 463--470.

\bibitem{bouguet2001opflow}
J.-Y. Bouguet, ``Pyramidal implementation of the affine lucas kanade feature
  tracker description of the algorithm,'' \emph{Intel Corporation}, vol.~5, pp.
  1--10, 2001.

\bibitem{lepetit2009epnp}
V.~Lepetit, F.~Moreno-Noguer, and P.~Fua, ``Epnp: An accurate o (n) solution to
  the pnp problem,'' \emph{International journal of computer vision}, vol.~81,
  no.~2, pp. 155--166, 2009.

\bibitem{tsai1989handeye}
R.~Y. Tsai and R.~K. Lenz, ``A new technique for fully autonomous and efficient
  3d robotics hand/eye calibration,'' \emph{Robotics and Automation, IEEE
  Transactions on}, vol.~5, no.~3, pp. 345--358, 1989.

\bibitem{hughes2015high}
M.~Hughes and G.-Z. Yang, ``High speed, line-scanning, fiber bundle
  fluorescence confocal endomicroscopy for improved mosaicking,''
  \emph{Biomedical optics express}, vol.~6, no.~4, pp. 1241--1252, 2015.

\bibitem{chang2013stereo}
P.-L. Chang, D.~Stoyanov, A.~J. Davison \emph{et~al.}, ``Real-time dense stereo
  reconstruction using convex optimisation with a cost-volume for image-guided
  robotic surgery,'' in \emph{Medical Image Computing and Computer-Assisted
  Intervention--MICCAI 2013}.\hskip 1em plus 0.5em minus 0.4em\relax Springer,
  2013, pp. 42--49.

\end{thebibliography}

\end{document}